\documentclass[journal]{IEEEtran}
\usepackage{amsmath,amsfonts}
\usepackage{algorithmic}
\usepackage{algorithm}
\usepackage{array}
\usepackage[caption=false,font=normalsize,labelfont=sf,textfont=sf]{subfig}
\usepackage{textcomp}
\usepackage{stfloats}
\usepackage{url}
\usepackage{verbatim}
\usepackage{graphicx}
\usepackage{cite}
\usepackage{multirow}
\usepackage{booktabs} 
\usepackage{diagbox}
\makeatletter

\begin{document}
	\title{CSHNet: A Novel Information Asymmetric Image Translation Method}

	\author{Xi Yang, \IEEEmembership{Senior Member,~IEEE},
            Haoyuan Shi,
            Zihan Wang,
			Nannan Wang, \IEEEmembership{Senior Member,~IEEE},
			and \\
   Xinbo Gao, \IEEEmembership{Fellow,~IEEE}
		\thanks{
%		 is with the State Key Laboratory of Integrated Services Networks, School of Telecommunications Engineering, Xidian University, Xi’an 710071, China (e-mail: duanss@stu.xidian.edu.cn).
		
		Xi Yang, Zihan Wang and Nannan Wang are with Xidian University, Xi’an 710071, China (e-mail: yangx@xidian.edu.cn;zihan\_w@stu.xidian.edu.cn; nnwang@xidian.edu.cn).

        Haoyuan Shi is with the Hangzhou Institute of Technology, Xidian University, Hangzhou 311231, China (e-mail: xdshy@stu.xidian.edu.cn).
		
		Xinbo Gao is with the Chongqing Key Laboratory of Image Cognition, Chongqing University of Posts and Telecommunications, Chongqing 400065, China, and also with the School of Electronic Engineering, Xidian University, Xi’an 710071, China (e-mail: gaoxb@cqupt.edu.cn; xbgao@mail.xidian.edu.cn).		
		}
	}

% {Shell \MakeLowercase{\textit{et al.}}: Bare Demo of IEEEtran.cls for IEEE Journals}


\maketitle
\begin{abstract}
Despite the considerable advancements in cross-domain image translation, a significant challenge remains in addressing information asymmetric translation tasks such as SAR-to-Optical and Sketch-to-Instance conversions. These tasks involve transforming data from a domain with limited information into one with more detailed and richer content. Traditional CNN-based methods, while effective at capturing intricate details, often struggle to grasp the overall structural composition of the image, leading to unintended blending or merging of distinct regions within the generated images. In light of these limitations, research has increasingly turned toward Transformers. Though Transformers excel at capturing global structures, they often lack the ability to preserve fine-grained details. Recognizing the importance of both detailed features and structural relationships in information asymmetric translation tasks, we introduce the CNN-Swin Hybrid Network (CSHNet). This network employs a novel bottleneck architecture featuring two key modules: Swin Embedded CNN (SEC) and CNN Embedded Swin (CES), which together form the SEC-CES-Bottleneck (SCB). Within this structure, SEC capitalizes on CNN's capability for detailed feature extraction while incorporating the Swin Transformer's inherent structural bias. In contrast, CES preserves the Swin Transformer's strength in maintaining global structural integrity, while compensating for CNN's tendency to emphasize detail. In addition to the SCB architecture, CSHNet integrates two essential components designed to improve cross-domain information retention and ensure structural consistency. The Interactive Guided Connection (IGC) fosters dynamic information exchange between SEC and CES, encouraging a deeper understanding of image details. At the same time, Adaptive Edge Perception Loss (AEPL) is implemented to preserve well-defined structural boundaries throughout the translation process. Experimental evaluations demonstrate that CSHNet surpasses current state-of-the-art methods, achieving superior results in both visualization and performance metrics across scene-level and instance-level datasets. Our code is available at: https://github.com/XduShi/CSHNet.

\end{abstract}
\begin{IEEEkeywords}
CNN, Swin Transformer, SEC-CES-Bottleneck (SCB), Interactive Guided Connection (IGC), Adaptive Edge Perception Loss (AEPL)
\end{IEEEkeywords}
\IEEEpeerreviewmaketitle

\begin{figure}[htb]
	\centering
	\includegraphics[width=1\columnwidth]{"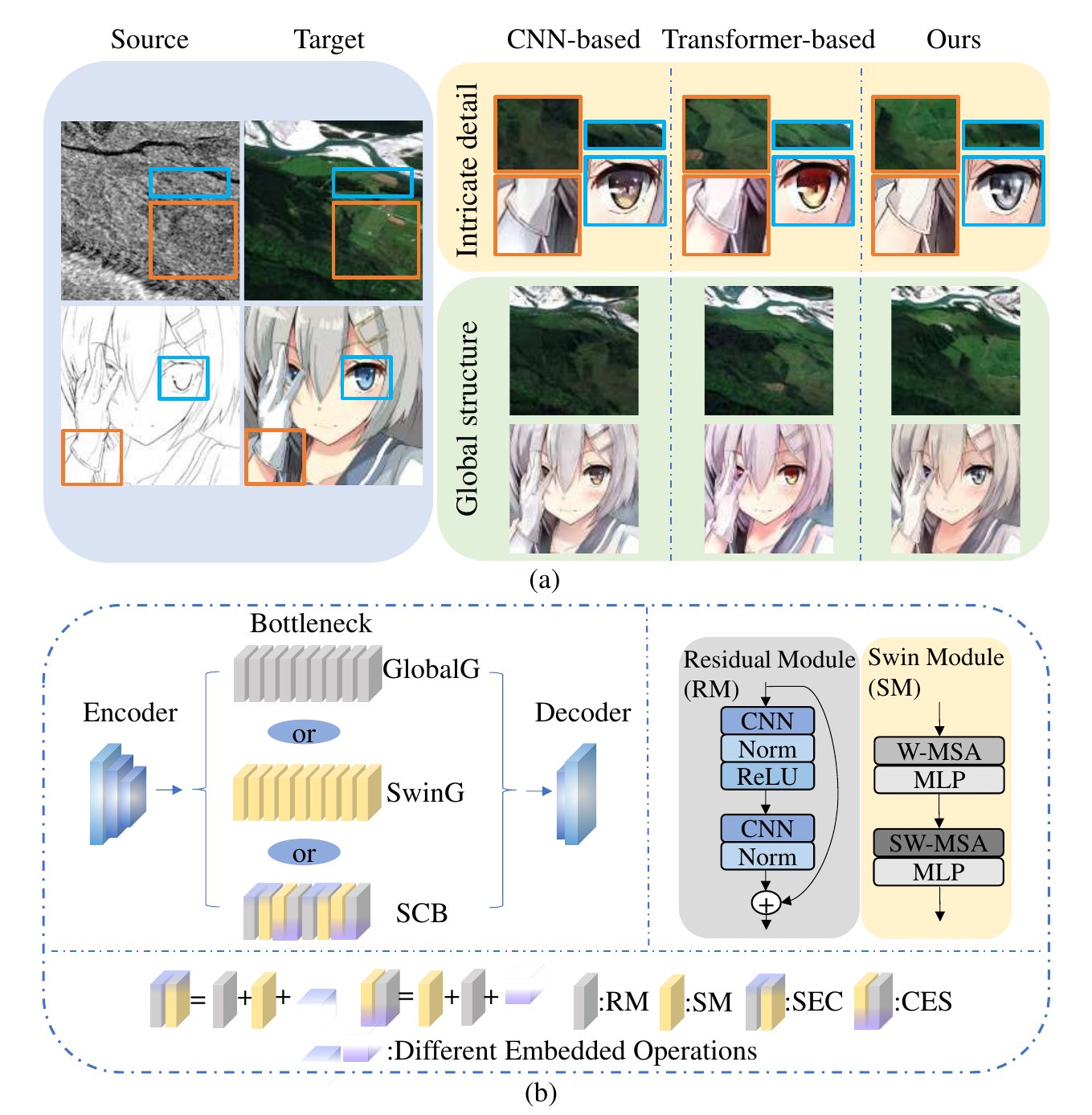"} 
	\caption{The network architectures adopted in existing methods and the associated challenges.red(a) Visualization of the network architectures for the methods described in (b). Detailed structural information is highlighted in the blue and orange boxes, representing the CNN-based bottleneck (GlobalG), Transformer-based bottleneck (SwinG), and the proposed SEC-CES-bottleneck. (b) Image-to-Image translation network based on Encoder-Bottleneck-Decoder paradigm. The proposed SCB has a similar hierarchical structure to GlobalG and SwinG but is not a simple stacking of RMs and SMs. It consists of SEC and CES cross-combination, considering both CNN and Transformer.}
	\label{fig1}
\end{figure}

\section{INTRODUCTION}

\IEEEPARstart{W}{ith} the development of deep learning, image translation tasks continue to evolve in various application fields. Among them, tasks such as style transfer \cite{chen2023controlstyle,xu2021drb,zhang2022exact,shen2023style} and makeup transfer \cite{chang2018pairedcyclegan,deng2021spatially,sun2022ssat, 10247199} have greatly enriched human visual experience. Additionally, image translation plays an essential role in the field of remote sensing, i.e., SAR-to-Optical (S2O) translation \cite{yang2022fg,yang2022sar} and Infrared-to-Optical (I2O) translation \cite{ozkanouglu2022infragan} address parts of SAR and infrared image recognition challenges.

While general image translation tasks have seen widespread application, image translation under conditions of information asymmetry, particularly from domains with deficient information to those with sufficient information, remains underdeveloped. As depicted in Figure \ref{fig1}(a), the source domain comprises scene-level images (SAR) lacking specific texture and structural information, and instance-level images (sketches) with ambiguous representational details (e.g., composition, characteristics). In contrast, the target domain includes a wealth of recognizable information. Consequently, data pairs exhibiting asymmetric information pose significant challenges for cross-domain image translation.

Most of the existing image translation methods are based on convolutional neural networks (CNNs) \cite{lecun1998gradient}, such as Pix2pix \cite{isola2017image}, Pix2pixHD \cite{wang2018high}, and CycleGAN \cite{zhu2017unpaired}. These methods typically consist of structures predominantly composed of 9 Residual Modules (RMs) \cite{isola2017image} and stacks of convolutional operations, exemplified by GlobalG \cite{wang2018high} in Figure \ref{fig1}(b). On the one hand, the local modeling property of CNN is prone to detailed bias, and ignores structural information. On the other hand, as the depth of the network increases, the repetitive convolutional feature extraction process may result in the loss of finer details. Furthermore, static receptive fields are inflexible to adapt to input content resulting in region blending. In the context of scene-level and instance-level data, structural information refers to the independence between different regions, wherein the network demonstrates accurate structural information if there is clear distinction and no cross-mixing between regions. However, CNN-based approaches struggle to ensure independence between regions, as indicated by the orange boxes of GlobalG in Figure \ref{fig1}(a). These inherent shortcomings hinder the intensive development of CNN-based image translation methods.

To address the local bias inherent in CNNs, visual Transformers have emerged as a viable alternative. Starting from ViT \cite{dosovitskiy2020image}, initially designed for simple image classification tasks, to Swin Transformer \cite{liu2021swin}, which serves as a versatile backbone for computer vision applications, Transformers have increasingly rivaled CNNs in various domains. Unlike the local receptive field of CNNs, Transformers leverage a self-attention mechanism to capture global interactions among contexts, with its structural bias well-established \cite{zhang2022delving}. Moreover, ViTGAN \cite{lee2021vitgan} and TransGAN \cite{jiang2021transgan} have demonstrated the image generation capabilities of Transformer. To establish a unified network structure, we introduce SwinG, based on the Swin Transformer, where the bottleneck consists of a simple stacking of 9 Swin Modules (SMs) as depicted in Figure \ref{fig1}(b). However, due to the absence of built-in inductive prior knowledge of interaction locality in Transformers, they often prioritize overall structural relationships at the expense of learning detailed information. This limitation can be observed in the generated images by SwinG, where certain detailed aspects, such as the highlights of the eyes, are lacking, as indicated in blue boxes of Figure \ref{fig1}(a). Hence, it becomes imperative to explore how to effectively combine the strengths of CNNs and Transformers to achieve superior performance in information asymmetric image translation tasks.

To address these challenges, our motivation lies in leveraging the complementary strengths of CNNs and Transformers. CNNs are particularly effective in capturing intricate local details due to their localized receptive fields, while Transformers excel in modeling global dependencies through their self-attention mechanism. Therefore, we propose a novel CNN-Swin Hybrid Network (CSHNet) tailored for information asymmetric datasets, aimed at mitigating the limitations of existing methods.  CSHNet reconstructs a novel bottleneck that combines detailed bias of CNN and structural bias of Transformer. As shown in Figure \ref{fig1}(b), SCB transcends a mere stacking of RMs and SMs, instead integrating a cross-combination of two key modules: Swin Embedded CNN (SEC) and CNN Embedded Swin (CES). SEC takes RM as a backbone and simultaneously embeds SM to introduce structural information. Conversely, CES relies on SM as its foundation and incorporates RM to supplement detailed information while preserving structural deviations.

Compared to pure CNN architectures, which excel at capturing local spatial details but suffer from limited global context modeling, CSHNet overcomes this limitation by integrating Swin Transformer modules to capture long-range dependencies and contextual relationships between distant regions. Conversely, pure Transformers are highly effective at modeling global interactions but often overlook fine-grained local details, especially in pixel-level feature representation. Transformers also suffer from increased computational complexity when processing high-resolution images at a fine granularity. CSHNet addresses these challenges by embedding CNN modules within Transformer layers, ensuring precise local detail extraction while maintaining structural consistency across regions. This hybrid design allows CSHNet to achieve a balance between local precision and global reasoning, outperforming pure CNNs in global context modeling and pure Transformers in fine-grained representation.

By harnessing the strengths of CNNs, CSHNet focuses on capturing fine-grained features to mitigate information loss during cross-domain translation. Simultaneously, it leverages the global perspective of Transformers to learn structural information between regions/instances, thereby ensuring coordination consistency. As depicted in Figure \ref{fig1}(a), images generated by SCB have complete details and clear structure, reducing information loss and region blending.
Moreover, since the semantic information in the feature maps becomes richer as the network deepens, to enhance the communication of low-level detail information and high-level semantic information, Interactive Guided Connection (IGC) is employed to guide feature map generation in the bottleneck using low-level information. Additionally, Adaptive Edge Perception Loss (AEPL) is introduced to further maintain structural consistency. AEPL calculates loss using dynamically threshold-optimized edge maps, thereby constraining the contour boundary information between real and generated images to mitigate the region blending phenomenon.

The main contributions of the paper are as follows:
\begin{itemize}
	\item We design a novel SEC-CES-Bottleneck to combine CNN and Transformer, which consists of two modules: Swin Embedded CNN and CNN Embedded Swin.
	\item To further enhance detail integrity, IGC is proposed to utilize low-level information to guide high-level semantic information within the bottleneck, thereby reducing cross-domain information loss.
	\item Furthermore, we introduce dynamic threshold-based AEPL to impose stricter constraints on the structural information of the generated images.
\end{itemize}

 \section{Related Work}
\subsection{GAN based on CNN}

Goodfellow et al. \cite{goodfellow2014generative} proposed GAN to achieve Noise-to-Image generation using adversarial training. To enhance network performance and stabilize the training process, DCGAN \cite{radford2015unsupervised} first presented the idea of combining GAN and CNN. Subsequently, CNN-based GANs and their variants have been extensively employed in computer vision tasks.

BEGAN \cite{berthelot2017began} used an autoencoder as a discriminator to improve the convergence speed without compromising the quality of generated images. BigGAN \cite{brock2018large} achieved remarkable results in class-conditional image synthesis by balancing fidelity and diversity through orthogonal regularization. In style transfer tasks, Pix2pix, pioneering work with a supervised approach, established the correspondence between source and target domain. Meanwhile, unsupervised style transfer methods \cite{ zhu2017unpaired,kim2017learning,yi2017dualgan,kim2019u,zheng2021one,yang2023scsp} have also made significant advancements. Li et al.\cite{li2019asymmetric} proposed an asymmetric GAN that learns additional information by introducing an auxiliary variable to perform unpaired image-to-image translation. IncrementalGAN \cite{9126831} tackled scalability issues in multi-domain settings by enabling incremental learning of new domains without retraining the entire model. A recent survey \cite{ning2023multi} comprehensively reviewed multi-view frontal face image generation methods, analyzing deep learning architectures and datasets. Furthermore, some models \cite{choi2018stargan,liu2021divco} achieved the translation between multiple styles. These approaches have facilitated the development of CNN-based GANs.

\subsection{GAN based on Vision Transformer}

The advent of ViT marked the dawn of the Transformer era in computer vision. Following ViT, numerous variants have emerged, each contributing to the advancement of this paradigm. Swin Transformer, for instance, reduced the computational complexity using the shift window mechanism and achieved satisfactory performance. To improve training stability, Swin v2 \cite{liu2022swin} incorporated strategies such as post-normalization, scaled cosine attention, and continuous relative position bias. In contrast, CSWin Transformer \cite{dong2022cswin} introduced a cross-window self-attentive mechanism. Xia et al. \cite{xia2022vision} proposed a Deformable Attention Transformer tailored for intensive prediction tasks.

The success of ViT has spurred further advancements in GANs. ViTGAN and TransGAN based on pure Transformer architectures have demonstrated the capability for image generation in unsupervised settings. Additionally, a high-resolution image generation network \cite{zhang2022styleswin} consisting of Transformer-based generators emerged. Moreover, ITTR \cite{zheng2022ittr} combined CNN and Transformer in the generator to accomplish unpaired Image-to-Image translation. Li et al.  \cite{li2023swin} proposed a Transformer-based Generative Adversarial Network model for high-resolution unpaired image translation. Meanwhile, SwinIT \cite{10399932} introduced a hierarchical Swin-transformer framework that eliminates the need for cycle consistency. In summary, the combination of Transformer and GAN exhibits promising potential for further development.

\subsection{Hybrid Networks Combining CNN and Transformer}
In CNN, convolutional operations excel at extracting local features but are limited in capturing global feature representations. Conversely, Vision Transformers leverage cascaded self-attention modules to capture feature dependencies over long distances but may overlook the details of local features. Recognizing the complementary strengths of these two approaches, several methods combining CNNs and Transformers have emerged.

Peng et al. \cite{peng2021conformer} proposed Conformer, the first dual backbone network combining CNN with Vision Transformer, aimed at enhancing feature representation learning. Mobile-former \cite{chen2022mobile} implemented MobileNet and Transformer in parallel, establishing a bidirectional bridge between them. In addition, CvT \cite{wu2021cvt} and CMT \cite{guo2022cmt} integrated convolutional layers into Vision Transformer to leverage the advantages of both architectures. Furthermore, a novel Transformer-style ``plug-and-play" module \cite{li2022contextual} was proposed to aggregate dynamic contextual information from the self-attentive mechanism in the Transformer with static contextual information from convolution operations. Different from these methods mainly used in classification, segmentation, detection and recognition tasks, Torbunov et al. \cite{torbunov2023uvcgan} and Kim et al. \cite{kim2022instaformer} applied CNN and Transformer hybrid networks to the image translation task. Additionally, hybrid networks have played an important role in the field of image inpainting \cite{phutke2023nested}, image enhancement \cite{yang2022rethinking} and person re-identification \cite{10196489}.

Hybrid CNN-Transformer networks combine the local detail extraction efficiency of pure CNNs with the global dependency modeling strength of pure Transformers \cite{Yang2022Hybrid} , addressing CNNs' limitations in long-range relationships and Transformers' challenges in fine-grained feature representation and computational efficiency \cite{Lee2024MetaFormer}. By integrating convolutional layers for local precision and self-attention mechanisms for global reasoning, hybrid architectures strike a balance between the strengths of pure CNNs and pure Transformers.

\subsection{Image-to-Image Translation}
Image translation aims at converting images of one given style to another given style, typically represented by Pix2pix and CycleGAN.

\emph{\textbf{Multimedia Applications:}} Zhang et al. \cite{zhang2017style} combined a residual U-net with an auxiliary classifier GAN to achieve style transfer for anime sketches. Akhtar et al. \cite{akhtar2022unsupervised} proposed a cartoonish style transfer method using multiple discriminators. Xie et al. \cite{xie2023unpaired} proposed an unsupervised image translation method based on shortest path regularization. Moreover, considering the high computational cost and memory usage of classical models, lightweight networks ASAPNet \cite{shaham2021spatially} and OMGD \cite{ren2021online} have emerged. Recently, Parmar et al. \cite{parmar2024one} proposed a method combining diffusion models and adversarial learning for both paired and unpaired image translation tasks. Recent advancements in flexible semantic image translation, such as FlexIT \cite{couairon2022flexit}, leverage multimodal embedding spaces to align image and text representations. Unsupervised image-to-image translation methods, such as the enhanced Swapping Autoencoder model \cite{shahfar2022unsupervised}, effectively disentangle structure and texture through gradient reversal layers.

\emph{\textbf{Remote Sensing Applications:}} Image-to-Image translation networks have found extensive applications in the field of remote sensing. Turnes et al. \cite{turnes2020atrous} proposed an atrous convolution-based CGAN architecture to implement S2O image translation. Subsequently, to solve complex scenes in optical images and high-frequency speckle noise in SAR images, a feature-guided S2O image translation method \cite{zhang2020feature} was presented. Unlike previous works, FG-GAN generated fine-grained optical images with an unsupervised approach. Furthermore, there have been notable contributions in visible-to-infrared and optical-to-SAR image translation. Han et al. Han et al. \cite{han2023aerial} introduced AVIID, the first dataset for aerial visible-to-infrared image translation. Han et al. \cite{han2024dr}  developed DR-AVIT, a GAN-based model using disentangled representation learning and dual geometry-consistency constraints. Ye et al. \cite{ye2022optical} proposed a multiscale masked structure feature method to improve optical-to-SAR image matching. These applications have addressed certain challenges in SAR and infrared image recognition problems, significantly advancing the field of remote sensing.

\begin{figure*}[t]
	\centering
	\includegraphics[width=1\textwidth]{"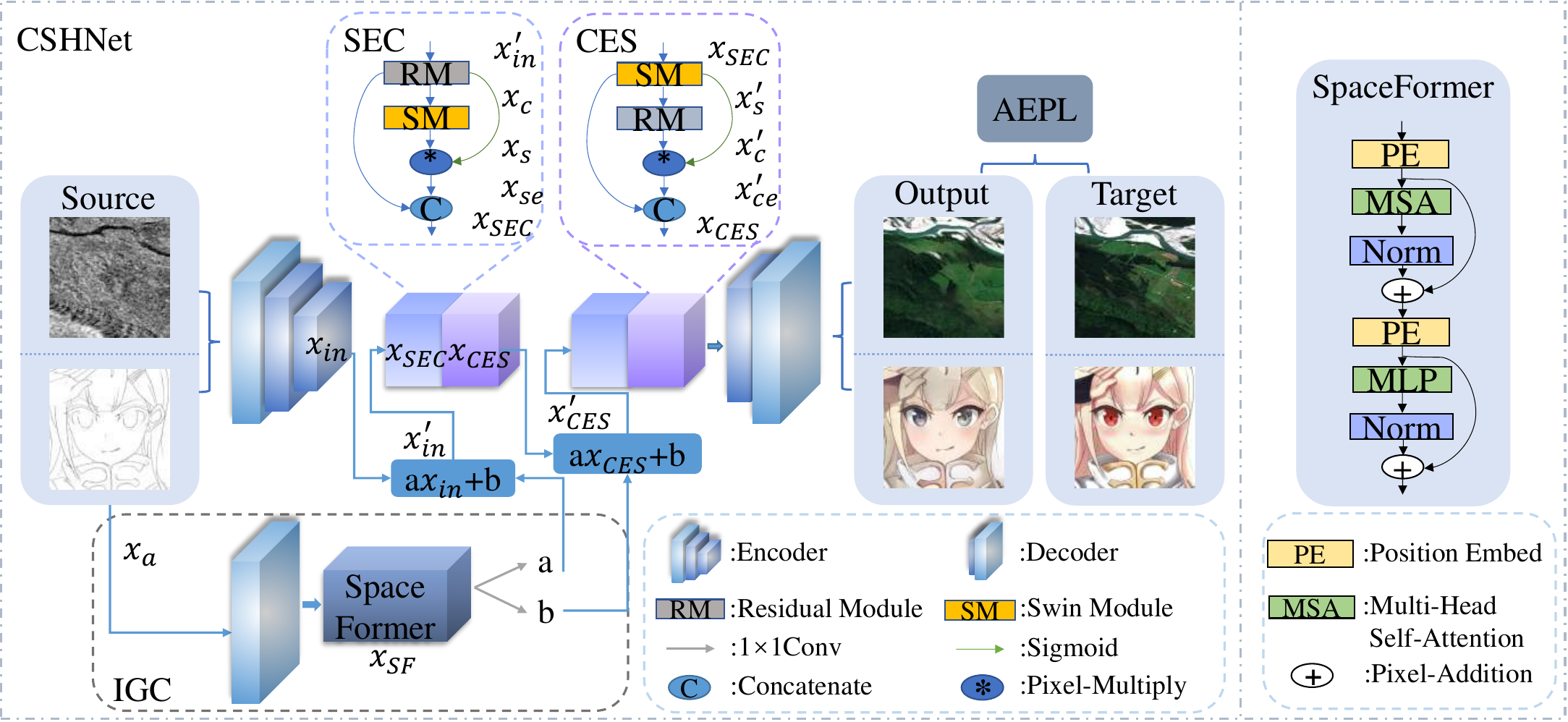 "}
	
	\caption{The proposed CSHNet framework for information asymmetric image translation. It contains three main components: SEC-CES-Bottleneck (SCB),  Interactive Guided Connection (IGC) and  Adaptive Edge Perception Loss (AEPL).}
	\label{fig2}
\end{figure*}

\section{Methodology}

Recently, image translation technology has been developing rapidly. Nonetheless, in scenarios where a substantial divergence in information content between two domains exists, such as the incorporation of semantic, textural, and color information during SAR-to-optical translation, conventional methods often encounter challenges in maintaining both detailed and structural fidelity in generated images. Therefore, particularly for tasks involving information asymmetry in image translation, we propose CSHNet, a hybrid network combining CNN and Swin Transformer. The overall structure is delineated in Figure \ref{fig2}. The encoder maps source domain images with relatively sparse information to latent space. The SEC-CES-Bottleneck (SCB) facilitates the extraction of detailed and structural information. Moreover, to compensate for the neglect of low-level features caused by network deepening, Interactive Guided Connection (IGC) operates on source domain images processed by relatively few network layers, extracting low-level information and generating learnable parameters ${a}$ and ${b}$ to augment the details of semantic features at bottlenecks.  Subsequently, the feature map undergoes transformation to the target domain via the decoder. To further bolster the learning of regional relationships and ensure alignment between real and generated images, we introduce Adaptive Edge Perception Loss (AEPL) into our framework.

\subsection{SEC-CES-Bottleneck (SCB)}

SCB amalgamates SEC and CES, with each module emphasizing detailed and structural biases, respectively, while also compensating for certain deficiencies present in the other. Recent advancements in encoder-decoder architectures, such as real-time 3D face alignment networks with efficient deconvolution layers \cite{ning2020real}, demonstrate effective feature fusion and computational efficiency. While existing approaches often alternate between convolutional layers and Transformer layers or replace Transformer feed-forward layers with convolutional representations \cite{guo2022cmt}, SCB instead adopts a synergistic fusion of these methods.

\textit{\textbf{Swin Embedded CNN (SEC):}} SEC integrates the detailedness of CNN with the structural information bias of Swin Transformer. As depicted in Figure \ref{fig2} , SEC comprises two primary modules: RM and SM, which represent Residual Module and Swin Module in Figure \ref{fig1}(b). Initially, the encoded feature map ${{x'_{in}}}$ is processed through RM to capture channel and spatial information, yielding ${x_c}$. Due to the convolution operation, ${x_c}$ is presumed to encapsulate extensive detailed information. Subsequently, SM operates on ${x_c}$ to derive ${x_s}$, imbued with structural information. To further refine the detailed information within ${x_s}$, a sigmoid function is employed to activate ${x_c}$ as the weighting factor for detailed information. The multiplication of the weight derived from ${x_c}$ with ${x_s}$ is employed to accentuate the intricate details within the feature maps, particularly those reflecting structural deviations, thereby resulting in the production of an enhanced feature map denoted as ${x_{se}}$. In essence, ${x_c}$ serves as a reference to align the data distribution of ${x_s}$ as closely as feasible to that of ${x_c}$. Concurrently, SM facilitates the acquisition of structural information by ${x_s}$. Thus, ${x_{se}}$ integrates detailed information emphasized by RM with structural information provided by SM. Finally, ${x_c}$ and ${x_{se}}$ are merged along the channel dimension to produce ${x_{SEC}}$, represented as follows:
\begin{equation}\label{equ1}
	{x_{SEC}} = Concat({x_c},SM(RM({x'_{in}}))*\sigma ({x_c})),
\end{equation}
where ${x_c}=RM({x'_{in}})$, ${x_s}=SM({x_c})$, ${x_{se}}={x_s}*\sigma ({x_c})$. $\sigma$ is the sigmoid function.

\textit{\textbf{CNN Embedded Swin (CES):}} In contrast to SEC, CES operates under the influence of the Swin Transformer, prioritizing the contour structure characteristics of images. As illustrated in Figure \ref{fig2}, ${x_{SEC}}$ initially undergoes processing through SM to yield ${x'_s}$. Given the intrinsic structural bias of the Transformer architecture, ${x'_s}$ is delineated as a feature map that primarily encapsulates structural information. However, its global perspective may potentially lead to the oversight of detailed texture information. To compensate for the missing detail information in  ${x'_s}$, RM is introduced to generate ${x'_c}$, incorporating CNN's detail bias. In alignment with the procedure employed in SEC, we leverage the activation values of ${x'_s}$ as weighting factors for structural information. Subsequently, ${x'_c}$ is multiplied by these weights to enhance the representation of structural contour information within the feature maps. Subsequently, the enhanced feature map ${x'_{ce}}$ is merged with ${x'_s}$ along the channel dimension to produce the final output ${x_{CES}}$ as follows:
\begin{equation}\label{equ2}
	{x_{CES}} = Concat({x'_s},RM(SM({x_{SEC}}))*\sigma ({x'_s})),
\end{equation}
where ${x'_s}=SM({x_{SEC}})$, ${x'_c}=RM({x'_s})$, ${x'_{ce}}={x'_c}*\sigma ({x'_s})$. $\sigma$ is the sigmoid function. In CES, we leverage the Swin Transformer to extract structural bias information and integrate local features from CNN.

\subsection{Interactive Guided Connection (IGC)}
In light of the inclusion of texture detail information in source domain images, and to enhance the integrity of generated images at a detailed level, we introduce IGC, which extracts features through a shallow network structure and abstracts this low-level feature information into two learnable parameters. These parameters are utilized to optimize the feature maps within the bottleneck, thereby facilitating the preservation and refinement of detailed integrity in the generated images.

Specifically, as shown in Figure \ref{fig2}, the input ${x_a}$ passes through a series of convolution operations to extract low-level marginal detail information. To mitigate information loss stemming from repeated convolutions, features are re-optimized by SpaceFormer instead of conventional convolution. We employ the post-normalization technique proposed by Swin v2 to stabilize the training process, incorporating residual connections after normalization within SpaceFormer. Then, two ${1\times1}$ convolution operations are performed to ${x_{SF}}$ obtained from SpaceFormer, enabling the learning of distinct parameters ${a}$ and ${b}$. Here, ${a}$ is utilized to modulate the weight of detailed information in feature maps, while ${b}$ is then applied to supplement the information of the whole feature map.
\begin{equation}\label{equ3}
	{x'_{f}} = a{x} + b,
\end{equation}
where ${x}$ represents the feature map to be optimized, namely the input image after passing through the encoder as ${x_{in}}$ and after the first set of SEC-CES as ${x_{CES}}$. ${x'_{f}}$ signifies the resulting optimized feature map.

In the ablation experiments, we examine the individual effects of ${a}$ and ${b}$. The findings reveal that employing the ${ax+b}$ formulation yields optimal outcomes. Meanwhile, the feature
maps before and after IGC action are visualized. The results indicate that the IGC module can be implemented to guide and complement the generation of high-level semantic feature map details in the bottleneck.

\begin{figure}[t]
	\centering
	\includegraphics[width=0.85\columnwidth]{"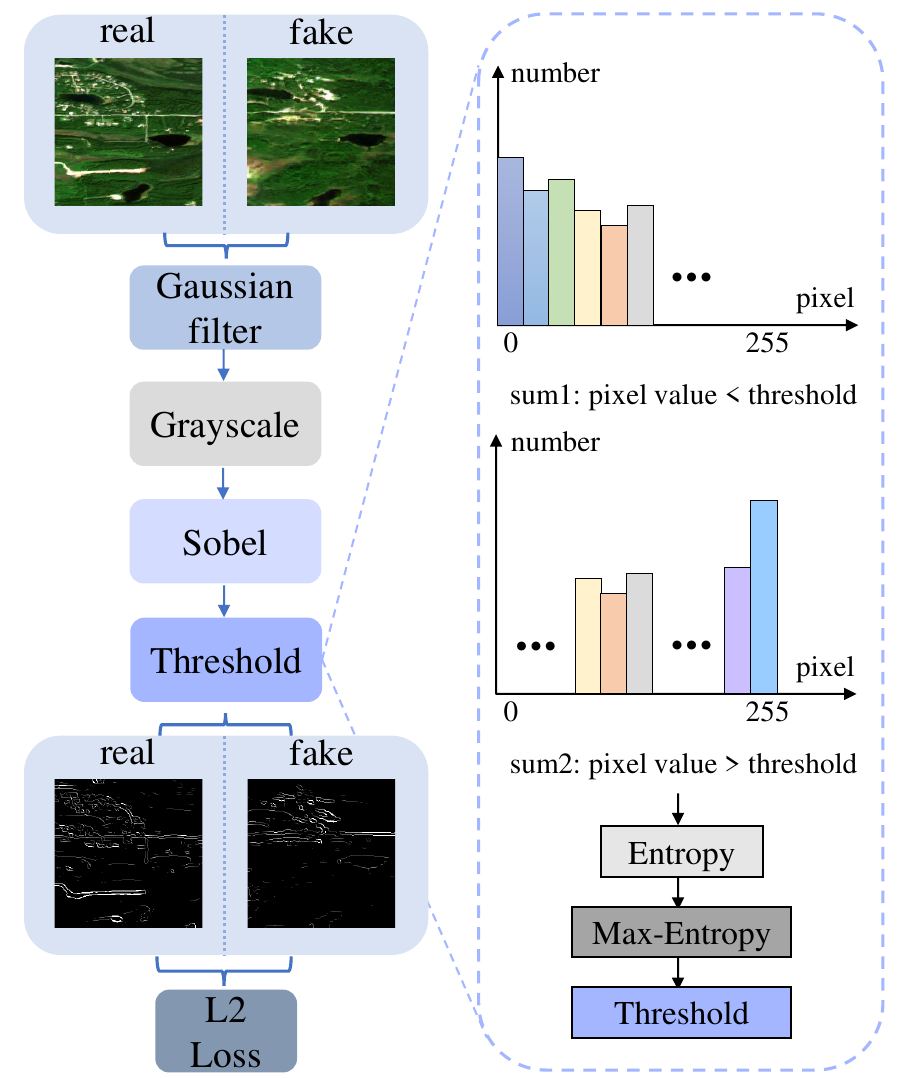"} 
	\caption{Flowchart of dynamic threshold-based AEPL.}
	\label{fig3}
\end{figure}

\begin{algorithm}[!htb]
    \renewcommand{\algorithmicrequire}{\textbf{Input:}}
    \renewcommand{\algorithmicensure}{\textbf{Output:}}
    \caption{Adaptive Edge Perception Loss (AEPL)}
    \begin{algorithmic}[1]
        \REQUIRE
            Real image $x_r$, Generated image $x_g$
        \ENSURE
            Adaptive Edge Perception Loss $\mathcal{L}_{AEPL}$
        
        \STATE \textbf{Step 1: Preprocessing}  
        Convert $x_r$ and $x_g$ to grayscale and apply Gaussian filtering to obtain smoothed images $x'_r$ and $x'_g$.
        
        \STATE \textbf{Step 2: Edge Detection}  
        Apply the Sobel operator to extract edge maps $f_{edge}(x'_r)$ and $f_{edge}(x'_g)$.
        
        \STATE \textbf{Step 3: Entropy-based Dynamic Thresholding}  
        \FOR{$q$ from $0$ to $255$}
            \STATE Calculate $H_1(q)$ using Equation \eqref{equ5}
            \STATE Calculate $H_2(q)$ using Equation \eqref{equ6}
            \STATE Compute total entropy $H(q)$ using Equation \eqref{equ7}
        \ENDFOR
        \STATE Find threshold $t$ that maximizes $H(q)$ using Equation \eqref{equ8}.
        
        \STATE \textbf{Step 4: Threshold Refinement}  
        Apply the threshold $t$ to emphasize primary contour structure:
        \[
        x_{edge} = f_{edge}(t, x'_r), \quad x_{edge} = f_{edge}(t, x'_g)
        \]
        
        \STATE \textbf{Step 5: Loss Calculation}  
        Compute the L2 loss between the refined edge maps:
        \[
        \mathcal{L}_{AEPL} = ||f_{edge}(t, x'_r) - f_{edge}(t, x'_g)||_2^2
        \]
        
        \RETURN $\mathcal{L}_{AEPL}$
    \end{algorithmic}
    \label{algor1}
\end{algorithm}
\subsection{Adaptive Edge Perception Loss (AEPL)}
From Figure \ref{fig1},  it is evident that existing methods applied to information asymmetric image translation often result in region blending in generated images. To achieve clearer separability of regions, we present dynamic threshold-based AEPL to constrain the region contour information of both real and generated images.

The detailed implementation process of AEPL is illustrated in Figure \ref{fig3} and Algorithm \ref{algor1}. To acquire edge region information, both real and generated images undergo grayscale conversion followed by Gaussian filtering to eliminate noise interference. Subsequently, the features of grayscale images are extracted using Sobel edge detector. In consideration of the potential blending of foreground and background information, a dynamic threshold is established to further refine the edge images. Specifically, an initial threshold value of ${q=0}$ is assigned, and the sums of pixels smaller than the threshold, ${p_l(q)}$, and larger than the threshold, ${p_h(q)}$, are computed. After that, the information entropy of different pixels is calculated using ${p_l(q)}$ and ${p_h(q)}$.
\begin{equation}\label{equ5}
	{H_1}(q) =  - \sum\limits_{i = 0}^q {\frac{{p(i)}}{{{p_l}(q)}}\log \frac{{p(i)}}{{{p_l}(q)}}} ,i \le q,
\end{equation}
\begin{equation}\label{equ6}
	{H_2}(q) =  - \sum\limits_{i = q + 1}^{255} {\frac{{p(i)}}{{{p_h}(q)}}\log \frac{{p(i)}}{{{p_h}(q)}}} ,i > q,
\end{equation}
\begin{equation}\label{equ7}
	{H(q)} = {H_1}(q) + {H_2}(q),
\end{equation}
where ${p(i)}$ denotes the count of pixels with a pixel value of ${i}$ across the entire image. The threshold ${t}$, which maximizes the information entropy ${H(q)}$, is determined through iterative computation.
\begin{equation}\label{equ8}
	t = q,when\;{H_{\max }}(q) = \max (H(q)),
\end{equation}
where ${H_{max}(q)}$ represents the maximum information entropy obtained at a threshold value of ${q}$. We define ${q}$ at this point as the dynamic threshold ${t}$ used to further optimize edge maps. Pixel values below ${t}$ typically contain background clutter, which can adversely affect the overall regional structure of the image. Therefore, we set pixel values below ${t}$ to ${0}$ for emphasizing the primary contour structure information within the edge map. ${x_{ini}}$ (the real image ${x_r}$ or the generated image ${x_g}$) is optimized using ${t}$ to get the final edge map ${x_{edge}}$. We adopt  ${L2}$ norm to quantify the similarity between the edge maps of real and generated images.
\begin{equation}\label{equ9}
	{\mathcal{L}_{AEPL}} = ||{f_{edge}}{(t,x'_r)} - {f_{edge}}(t,x'_g)||_2^2,
\end{equation}
where ${f_{edge}}\left(  \cdot  \right)$ denotes the edge map processing function with threshold constraint. ${x'_r}$ and ${x'_g}$ indicate the real image and the generated image, respectively, after undergoing Gaussian filtering, grayscale conversion, and Sobel operator processing.

Furthermore, CSHNet incorporates adversarial loss (${\mathcal{L}_{GAN}}$), feature matching loss (${\mathcal{L}_{feat}}$), and content loss (${\mathcal{L}_{cont}}$) proposed in Pix2pixHD to minimize the distribution gap between real and generated images. Therefore, the full objective function is:
\begin{equation}\label{equ10}
	\begin{split}
		{\mathcal{L}_{total}} & = {\lambda _{GAN}}*{\mathcal{L}_{GAN}} + {\lambda _{feat}}*{\mathcal{L}_{feat}} \\
		& + {\lambda _{cont}}*{\mathcal{L}_{cont}} + {\lambda _{AEPL}}*{\mathcal{L}_{AEPL}},
	\end{split}
\end{equation}
where ${\lambda _{GAN}}$, ${\lambda _{feat}}$, ${\lambda _{cont}}$, and ${\lambda _{AEPL}}$ are hyper-parameters employed to balance different parts of training loss.

\section{\textbf{Experiments}}

\subsection{Experimental Settings}
%\textbf{1. Baseline Model}

\textit{\textbf{Baseline:}} For fairness, we adopt GlobalG in Pix2pixHD \cite{wang2018high} as a CNN-based comparison method. Additionally, we design SwinG, a network architecture based on Swin Transformer, as a Transformer-based comparative experiment, as depicted in Figure \ref{fig1}(b). Meanwhile, we compare CSHNet with SOTAs in recent years, including Pix2pix, CycleGAN, DivCo \cite{liu2021divco}, ASAPNet, SRC \cite{jung2022exploring}, Santa \cite{xie2023unpaired}, UVCGAN \cite{torbunov2023uvcgan} StegoGAN \cite{wu2024stegogan} and ATME \cite{solano2023look}. Pix2pix and CycleGAN address paired and unpaired image-to-image translation, respectively, while DivCo enhances generative diversity, ASAPNet optimizes computational efficiency, and SRC preserves semantic correspondence. Santa focuses on fine-grained representation, UVCGAN ensures unsupervised consistency, StegoGAN integrates steganographic techniques, and ATME employs attention mechanisms for refined detail capture. Collectively, these methods tackle dataset pairing, output variability, and detail fidelity, providing a foundation upon which CSHNet advances performance and applicability. All methods are implemented using publicly available codebases and executed under the same experimental environment to ensure a fair comparison.

%add(R.J1My-1) “More datasets." 
\textit{\textbf{Datasets:}} Based on user needs, we select two
common datasets from remote sensing and multimedia applications for training and testing, including a scene-level dataset (SEN12 \cite{schmitt2018sen1}) and an instance-level dataset (Sketch2Anime). SEN12 dataset consists of 282,384 SAR-Optical image pairs acquired by Sentinel-1 and Sentinel-2. The training and test images are sourced from the Spring-Summer folder within the SEN12 dataset. We further collect a Sketch2Anime dataset in which anime images are from Danbooru2018 \cite{danbooru2018} and the corresponding sketches are generated by \cite{Anime2Sketch}. All anime images and sketches are aligned,  resulting in a total of 135,509 image pairs. To mitigate the impact of training time and GPU memory on experimental efficiency, we randomly and evenly select 3,000 pairs of images for training and 800 pairs for testing from each dataset. The distribution of training and test data is similar but non-overlapping.

\textit{\textbf{Implementation:}} Experiments are conducted on a computer with a single GTX TITAN with 12-GB GPU memory, and implemented on PyTorch framework. We utilize the Adam optimizer with control redistribution, the exponential decay rate of gradient squared and batch size set to 0.5, 0.999, and 1, respectively. 200 epochs are trained on all datasets to ensure model convergence. We follow the setup of Pix2pixHD using LSGAN and PatchGAN in our model. Furthermore, we set ${\lambda _{GAN}}$, ${\lambda _{feat}}$, ${\lambda _{cont}}$, and ${\lambda _{AEPL}}$ to 1, 10, 10 and 1 respectively to balance the training process of CSHNet.

\textit{\textbf{Evaluation Metrics:}} Six metrics are used to evaluate the quality of generated images, comprising objective metrics (PSNR, SSIM \cite{wang2004image}, FID \cite{heusel2017gans}, LPIPS \cite{zhang2018unreasonable} and RMSE) and a subjective indicator (NIQE \cite{mittal2012making}). PSNR quantifies image quality based on error-sensitive evaluation. SSIM assesses the similarity of images in terms of brightness, contrast, and structure. Higher PSNR and SSIM values generally correspond to better-quality images. FID measures the diversity of image generation by feeding images into a pre-trained InceptionV3 to extract feature distributions and calculate similarity. NIQE evaluates image quality by computing the distance between image feature model parameters and pre-established model parameters. Lower values of FID and NIQE indicate smaller differences between generated and real images. LPIPS quantifies perceptual similarity using deep neural networks, with lower values indicating greater visual similarity to real images. RMSE measures the average squared pixel differences, with lower values indicating better alignment with real images.
\begin{figure}[t]
	\centering
	\includegraphics[width=1\columnwidth]{"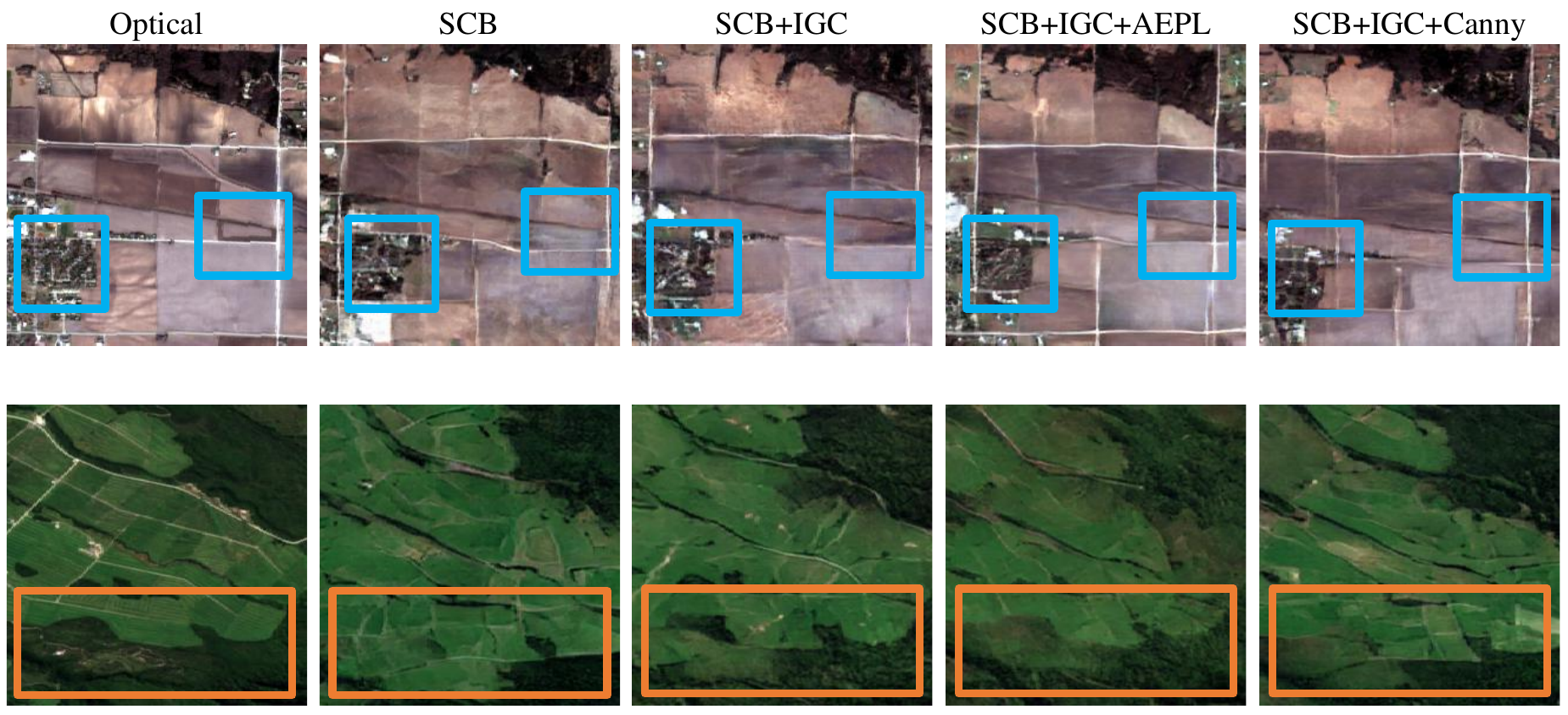"} % Reduce the figure size so that it is slightly narrower than the column. Don't use precise values for figure width.This setup will avoid overfull boxes.
	\caption{Visualization results of ablation experiments with different components on SEN12 dataset.}
	\label{fig4}
\end{figure}

\begin{figure}[t]
	\centering
	\includegraphics[width=1\columnwidth]{"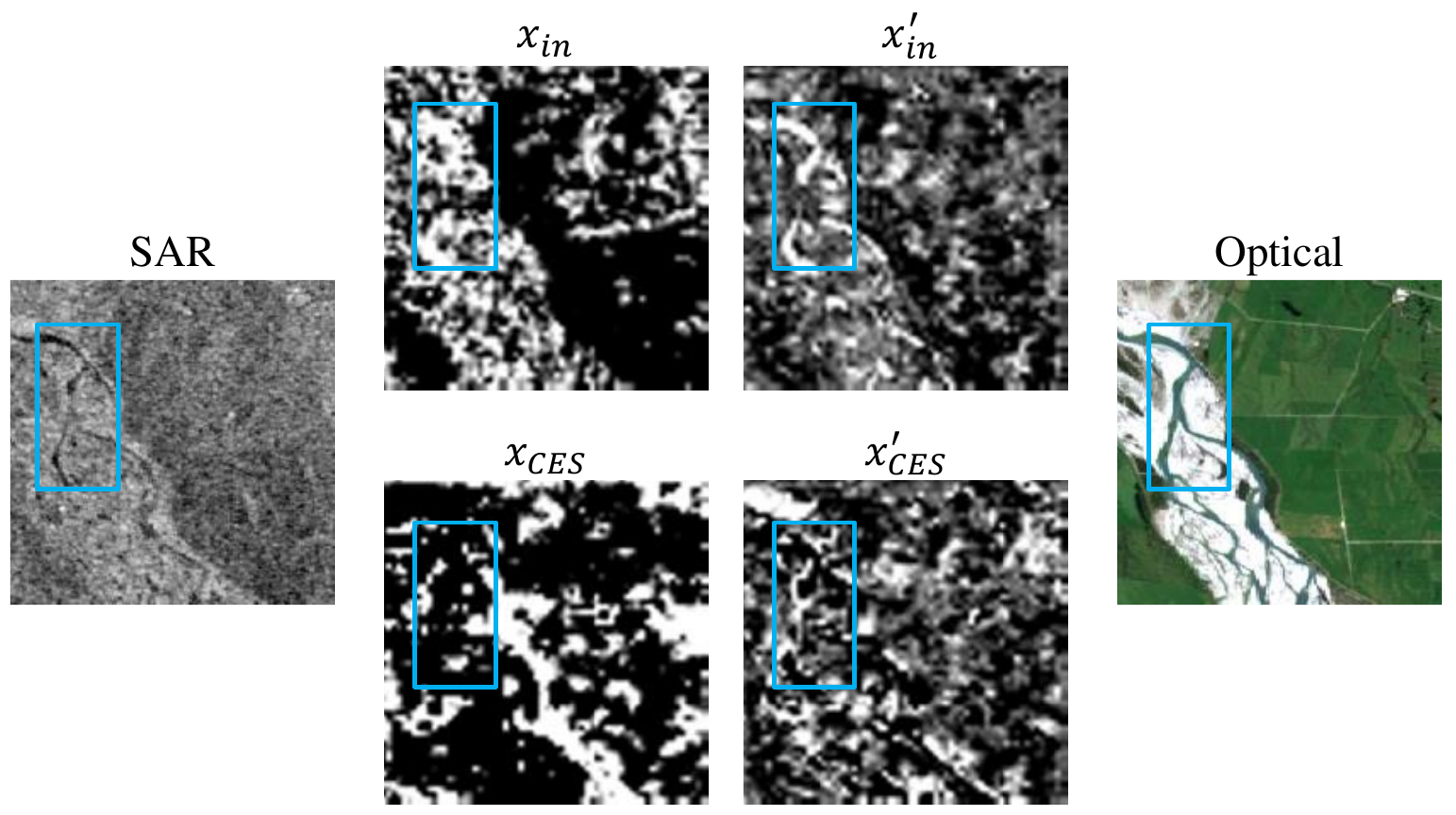"} 
	\caption{Visualization results of the feature maps before and after the action
of IGC. $x_i$ and $x_{CES}$ are the results before the optimization of parameters
$a$ and $b$. $x_{i}^{\prime}$ and $x_{CES}^{\prime}$ are the feature maps after IGC.}
	\label{fig5}
\end{figure}

\begin{table}[t]
	\centering
	\caption{Different implementations of bottleneck and IGC in CSHNet on SEN12 dataset.}
        \vspace{0.3mm}
	\begin{tabular}{lc|cccc}
		\toprule
		\multicolumn{2}{c|}{Component}                          & PSNR  $\uparrow$            & SSIM  $\uparrow$            & NIQE $\downarrow$         & FID  $\downarrow$           \\ \hline
		\multicolumn{1}{c|}{\multirow{4}{*}{Bottleneck}} & 4SEC & 14.97          & 0.2156          & 35.7          & 94.1          \\
		\multicolumn{1}{c|}{}                            & 4CES & \textit{15.03} & \textit{0.2156} & \textit{34.5} & 92.9          \\
		\multicolumn{1}{c|}{}                            & CSB  & \textbf{15.04} & 0.2148          & 36.0          & 102.6         \\
		\multicolumn{1}{c|}{}                            & SCB  & 14.98          & \textbf{0.2168} & \textbf{34.3} & \textbf{91.0} \\ \hline
		\multicolumn{1}{l|}{\multirow{3}{*}{IGC}}        & ax   & \textit{15.20} & 0.2153          & \textit{34.3} & \textbf{89.1} \\
		\multicolumn{1}{l|}{}                            & x+b  & 15.10          & \textit{0.2184} & 34.4          & 90.4          \\
		\multicolumn{1}{l|}{}                            & ax+b & \textbf{15.24} & \textbf{0.2477} & \textbf{34.1} & \textit{90.4} \\
		\bottomrule
	\end{tabular}
	\label{tab1}
\end{table}

\begin{figure*}[t]
	\centering
	\includegraphics[width=0.99\textwidth]{"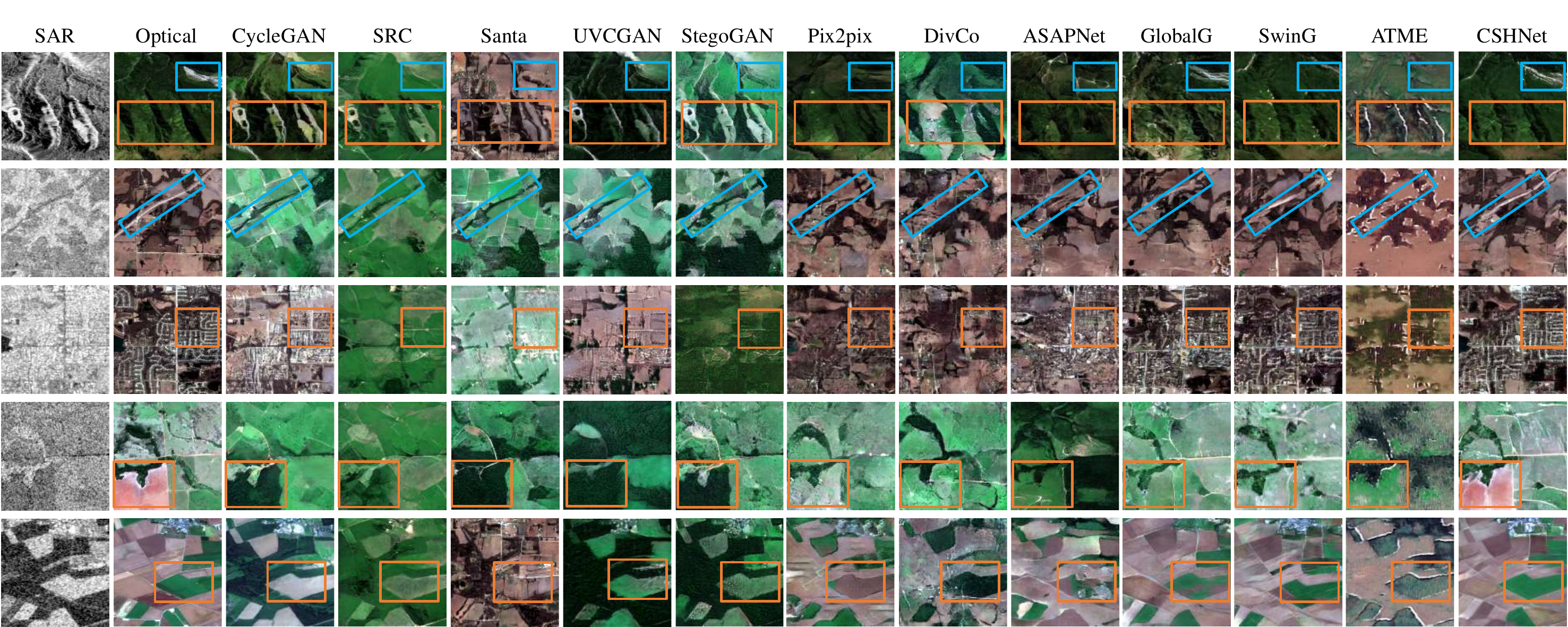"} 
	\caption{Visualization results of different methods for information asymmetric image translation on SEN12 dataset (scene-level).}
	\label{fig6}
\end{figure*}
\begin{figure*}
	\centering
	\includegraphics[width=0.99\textwidth]{"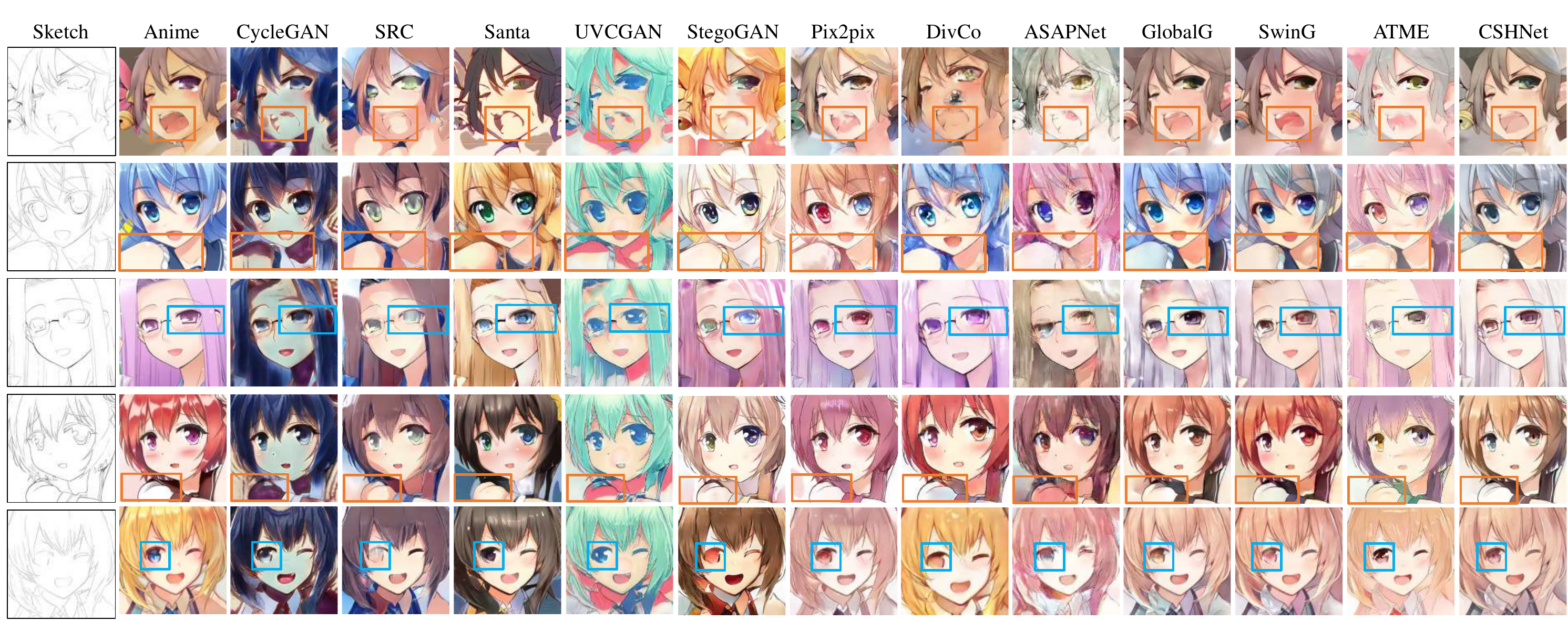"} 
	\caption{Visualization results of different models for information asymmetric image translation on Sketch2Anime dataset (instance-level).}
	\label{fig7}
\end{figure*}

\subsection{Ablation Study}
Given that optical images in SEN12 dataset encompass a significant amount of detailed texture information as well as distinct regional contour structures, all ablation experiments are conducted exclusively on SEN12 dataset.

Firstly, we compare different implementations of the bottleneck in CSHNet. As depicted in Table \ref{tab1}, 4SEC and 4CES denote the utilization of 4 SECs and 4 CESs in the bottleneck, i.e., focusing on either CNN or Swin Transformer exclusively. CSB indicates that the network adopts the structure of CES-SEC, contrary to the order employed by the proposed method SCB. ${2}$ SECs and ${2}$ CESs are applied to the proposed SCB with SEC-CES order, thereby considering both CNN and Swin Transformer aspects. \textbf{Bold} indicates the optimal and \textit{italic} signifies sub-optimal. Among the four evaluation metrics, SCB, which combines detailed and structural bias, achieves the best results. Since CSB prioritizes the overall structure, it leads to the loss of shallow detail information extracted by the network as its depth increases.  In contrast, SCB concentrates on processing detail information first, ensuring the richness of details in generated images.  Based on SCB, we conduct various forms of ablation experiments for IGC, namely ${ax}$, ${x+b}$, and ${ax+b}$. Table \ref{tab1} illustrates that ${ax+b}$ yields favorable scores across the majority of evaluation metrics, which facilitates detail integrity. Consequently, we abstract IGC as the form of ${ax+b}$.

\begin{table}
	\normalsize
	\caption{Different components in CSHNet on SEN12 dataset.}
        \vspace{0.3mm}
	\centering
	%\vspace{0.2cm}
	\setlength{\tabcolsep}{3pt}{
		\begin{tabular}{cccc|cccc}
			\toprule
			\multicolumn{4}{c}{Component} & \multicolumn{4}{c}{Metric} \\ \hline
			\begin{tabular}[c]{@{}c@{}}SCB \end{tabular}
			& \begin{tabular}[c]{@{}c@{}}IGC \end{tabular}
			& \begin{tabular}[c]{@{}c@{}}Canny \end{tabular}
			&  \begin{tabular}[c]{@{}c@{}}AEPL \end{tabular}
			&\begin{tabular}[c]{@{}c@{}}PSNR $\uparrow$\end{tabular}
			& \begin{tabular}[c]{@{}c@{}}SSIM $\uparrow$\end{tabular}
			& \begin{tabular}[c]{@{}c@{}}NIQE $\downarrow$\end{tabular}
			& \begin{tabular}[c]{@{}c@{}}FID $\downarrow$\end{tabular} \\ \hline
			\textbf{$\surd$} & $\times$ 		& $\times$ 			& $\times$ 			
			& 14.98 & 0.2168 & 34.3	&91.0 \\
			\textbf{$\surd$} & \textbf{$\surd$} & $\times$ 			& $\times$ 			
			& 15.24 & \textit{0.2477} & 34.1	& \textit{90.4} \\
   			\textbf{$\surd$} & $\times$ 		& \textbf{$\surd$} 			& $\times$ 			
			& 15.01  & 0.2142 & 33.9 	& 93.5\\ 
			\textbf{$\surd$} & $\times$ & $\times$ 			& \textbf{$\surd$} 			
			& 15.07 & 0.2256  & \textbf{33.7}	& 90.8 \\ 
			\textbf{$\surd$} & \textbf{$\surd$} & \textbf{$\surd$}  & $\times$ 			
			& \textit{15.26} & 0.2176 & 35.0 	& \textbf{88.2}\\
			\hline
			\textbf{$\surd$} & \textbf{$\surd$} & $\times$			& \textbf{$\surd$}
			& \textbf{16.37} & \textbf{0.2895} & \textit{33.9} 	& 90.5\\
			\bottomrule
	\end{tabular}}
	\label{tab2}
\end{table}

\begin{table*}[t] 
	\centering
        \setlength\tabcolsep{12pt}
	\caption{Evaluation metrics with different methods on SEN12 dataset (SAR-to-Optical). CycleGAN, SRC, Santa, UVCGAN and StegoGAN are unsupervised methods, others are supervised.}
	\small
	\begin{tabular}{c|c|cccccc}
		\toprule
		\diagbox{Method}{Metric}	&\makebox[0.08\textwidth][c]{Publication}       & PSNR $\uparrow$             & SSIM $\uparrow$             & NIQE  $\downarrow$        & FID $\downarrow$     & LPIPS $\downarrow$  &RMSE  $\downarrow$    \\ \hline
		CycleGAN	&\textit{CVPR17}     & 11.29          & 0.1492          & 36.1         & 109.5         & 0.653 & 65.2 \\
		SRC		&\textit{CVPR22}          & 12.05          & 0.1814          & 37.6          & 143.5         & 0.693 & 68.5 \\
		Santa	&\textit{CVPR23}		 & 10.34		  & 0.1134			& 32.5 & 92.1			& 0.688 & 72.3 \\
		UVCGAN	&\textit{WACV23}		 & 10.77		  & 0.1322			& \textbf{31.5} &133.6			& \textit{0.646} & 70.1 \\
        StegoGAN	&\textit{CVPR24}		 & 11.29		  & 0.1189		& 40.9 &131.3 & 0.675 & 68.3		\\
		\hline
		Pix2pix		&\textit{CVPR17}      & \textit{15.84} & \textit{0.2568} & 37.6          & 139.1         & 0.650 & 63.8 \\
		DivCo		&\textit{CVPR21}        & 12.00          & 0.1826          & 35.9          & 132.1         & 0.690 & 67.0 \\
		ASAPNet		&\textit{CVPR21}      & 14.51          & 0.2116          & \textit{32.1} 			& 107.9         & 0.663 & 66.2 \\
        ATME        &\textit{CVPR23}      & 14.51             & 0.2353         & 39.5          & 195.1  & 0.692 & 70.0 \\
		GlobalG		&\textit{CVPR18}      & 15.21          & 0.2168          & 35.2          & 103.5         & 0.693 & 64.3 \\
		SwinG		&\textit{CVPR21}        & 15.09          & 0.2169          & 34.6          & \textit{91.6} & 0.670 & \textit{61.7} \\
		CSHNet		&ours & \textbf{16.37} & \textbf{0.2895} & 33.9 & \textbf{90.5} & \textbf{0.635} & \textbf{45.1} \\
		\bottomrule
	\end{tabular}
	\label{tab3}
\end{table*}

\begin{table*}[ht] 
	\centering
        \setlength\tabcolsep{12pt}
	\caption{Evaluation metrics with different methods on the Sketch2Anime dataset (Sketch-to-Anime).}
	\small
	\begin{tabular}{c|c|cccccc}
		\toprule
		\diagbox{Method}{Metric}	&\makebox[0.08\textwidth][c]{Publication}       & PSNR $\uparrow$             & SSIM $\uparrow$             & NIQE  $\downarrow$        & FID $\downarrow$    & LPIPS $\downarrow$     & RMSE $\downarrow$   \\ \hline
		CycleGAN	&\textit{CVPR17}     & 9.43           & 0.4123          & 18.1         & 77.6          & 0.695 & 72.2 \\
		SRC		   &\textit{CVPR22}       & 11.86          & 0.5690          & 13.5         & 60.7          & 0.683 & 68.7 \\
		Santa	&\textit{CVPR23}	 & 10.93		  & 0.4891			& 29.9		   & 43.6		& 0.665 & 69.9 \\
		UVCGAN		&\textit{WACV23}       & 10.64          & 0.5179          & 11.4         & 104.3			& \textbf{0.651} & 71.0 \\
        StegoGAN	&\textit{CVPR24}		 & 11.08	  & 0.5111		& 33.9 &108.8		& 0.671 & 71.4 \\
		\hline
		Pix2pix		&\textit{CVPR17}      & 14.71          & 0.6205          & 14.6         & 64.8          & 0.696 & 66.8 \\
		DivCo	&\textit{CVPR21}        & 16.22          & 0.6430          & 14.9         & 72.1          & 0.683 & 68.3 \\
		ASAPNet		&\textit{CVPR21}      & 12.31          & 0.5024          & 10.9         & 77.2          & 0.663 & 69.5 \\
        ATME       &\textit{CVPR23}      & 14.77            & 0.6326        & 12.0          & 47.4   & 0.663 & 41.5       \\
		GlobalG		&\textit{CVPR18}      & 18.86          & 0.7717          & 9.6          & 43.7          & 0.676 & 40.2 \\
		SwinG	   &\textit{CVPR21}     & \textit{19.00} & \textit{0.7787} & \textit{9.2} & \textbf{41.1} & 0.669 & \textit{39.6} \\
		CSHNet		&ours  & \textbf{19.01} & \textbf{0.7820} & \textbf{9.0} & \textit{41.8} & \textit{0.661} & \textbf{23.9}\\
		\bottomrule
	\end{tabular}
	\label{tab4}
\end{table*}

After determining the structure of the bottleneck and IGC, we evaluate each component in CSHNet, i.e., SCB, IGC, and AEPL. The results are presented in Table \ref{tab2}. Upon incorporating IGC, all metrics demonstrate improvement, suggesting that IGC may enhance the texture fineness of generated images (blue boxes in Figure \ref{fig4}).  Meanwhile, the feature maps (i.e., $x_i$, $x_i^{\prime}$, $x_{CES}$ and $x_{CES}^{\prime}$) before and after the effect of IGC are visualized. As shown in the blue boxes in Figure \ref{fig5}, after the optimization of parameters $a$ and $b$, some low-level detail information of the feature map is enhanced, such as edges and corners. It indicates that the communication between the feature maps in the bottleneck and the low-level information extracted by the IGC is enhanced. With the incorporation of Canny edge detection, a modest decline in SSIM and FID is observed. However, this method demonstrates a enhancement in PSNR and NIQE. Following the introduction of AEPL, both PSNR and SSIM experience a marked increase, with SSIM improving from 0.2168 to 0.2256. Additionally, when IGC is implemented in conjunction with AEPL, both PSNR and SSIM show substantial gains, with SSIM rising significantly from 0.2477 to 0.2895. Orange boxes in Figure \ref{fig4} show that AEPL further enhances the structural consistency between real and generated images. To underscore the advantages of adaptive edge perception, we compare AEPL with Canny edge detection \cite{ding2001canny}, which also incorporates a thresholding operation. As demonstrated in Table \ref{tab2} and Figure \ref{fig4}, AEPL outperforms Canny edge detection operator. Since Canny edge detection operator only allows for setting a fixed threshold initially, AEPL adaptively adjusts the threshold based on the information entropy of images, rendering it more suitable for scene-level datasets with complex information.

\subsection{Comparison with State-of-the-Arts}
The experiments are conducted using two datasets: the scene-level dataset SEN12 and the instance-level dataset Sketch2Anime. We perform comparisons and analyses of visualization results and evaluation metrics across different Image-to-Image translation methods.

\textit{\textbf{Visualization Comparison:}} As depicted in Figure \ref{fig6}, the SEN12 dataset comprises a wealth of texture details and region structure information. Supervised methods (e.g., Pix2pix, DivCo, ASAPNet, ATME, GlobalG and SwinG) establish accurate scene-level mappings between SAR and Optical data. However, noticeable detail loss is observed in the generated images, such as the absence of white rivers and lines highlighted in blue boxes in rows 1 and 2. In addition, the regional contours are heavily blended, as evidenced by the orange boxes of Figure \ref{fig6}. Due to the structural bias of Transformer, SwinG generates images with improved regional contours compared to other SOTAs, as observed in rows 1 and 3 of Figure \ref{fig6}. On the other hand, unsupervised methods, such as CycleGAN, SRC, Santa, UVCGAN and StegoGAN, perform inadequately in the SAR-to-Optical translation task. They fail to ensure the correctness of the overall scene style and suffer from severe detail loss in the generated images. In contrast, the proposed CSHNet, employing supervised adversarial training, mitigates these shortcomings. Since it inherits both detail and structure biases, generated images from CSHNet maintain the integrity of texture details and the clarity of region contours. Similar results are observed in the Sketch2Anime dataset. As indicated by the orange boxes in the first row of Figure \ref{fig7}, CSHNet accurately captures the mouth contour, whereas other methods fail. The detail bias is further exemplified by the blue boxes in rows 3 and 5 of Figure \ref{fig7}, demonstrating CSHNet's attention to minute details such as eyes.

\begin{figure}
	\centering
	\includegraphics[width=0.5\textwidth]{"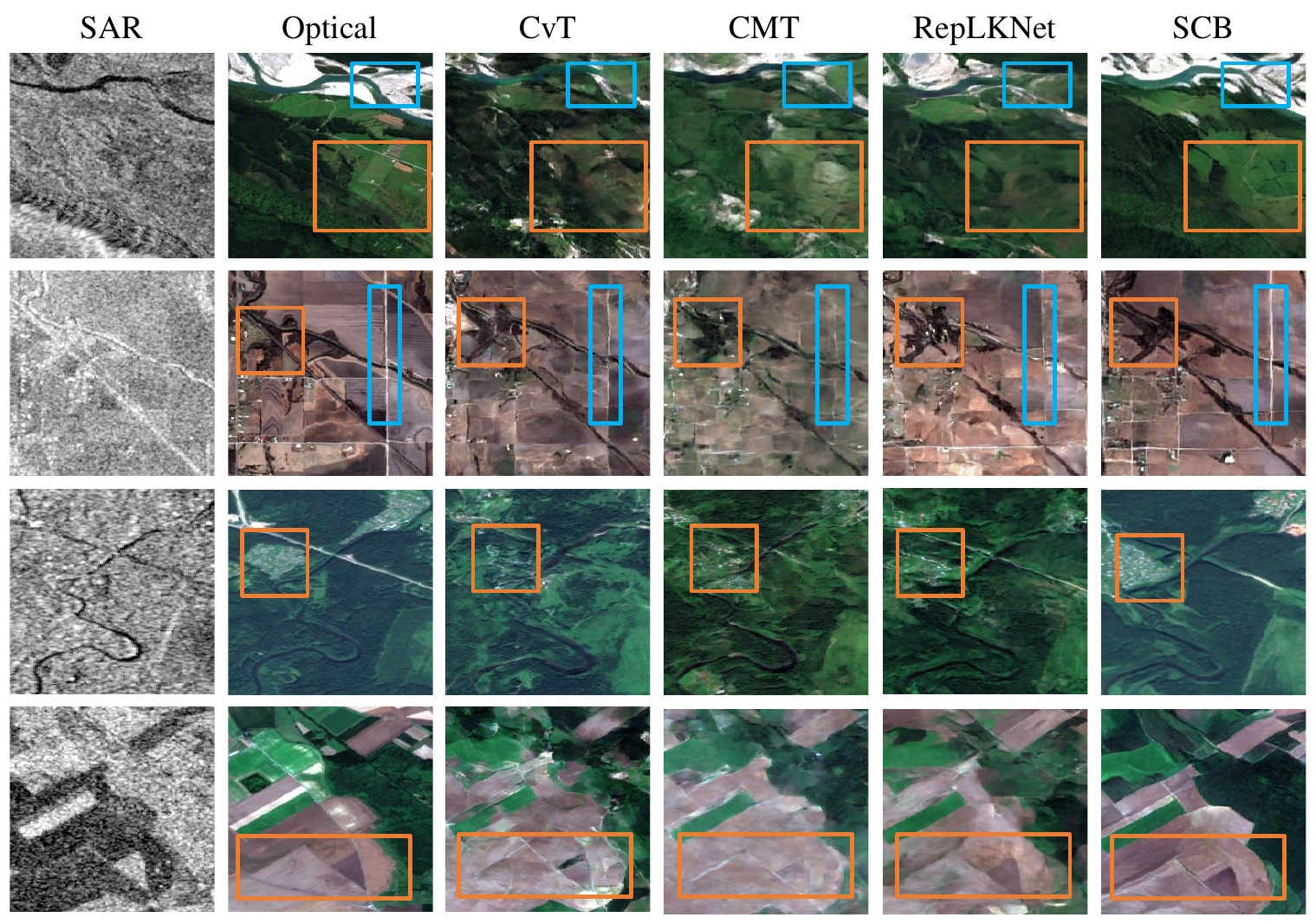"} 
	\caption{Visualization results of different CNN and Transformer hybrid networks (CvT, CMT, and SCB) and a network combining different convolutional kernel sizes (RepLKNet) on the SEN12 dataset.}
	\label{fig8}
\end{figure}
\textit{\textbf{Metrics Comparison:}}

Table \ref{tab3} and Table \ref{tab4} present the evaluation results of CSHNet alongside other SOTA methods on the SEN12 and Sketch2Anime datasets, respectively. On the SEN12 dataset, CSHNet excels in SSIM with a score of 0.2895, significantly outperforming other methods, demonstrating strong structural consistency. CSHNet also achieves the best LPIPS score of 0.635, indicating superior perceptual quality and visual similarity to real images compared to other methods. While NIQE slightly lags behind UVCGAN, the difference is minimal, suggesting room for improvement in statistical smoothness despite high overall fidelity. RMSE also stands out for CSHNet at 45.1, showing excellent pixel-level accuracy.

In Table \ref{tab4}, on the Sketch2Anime dataset, CSHNet achieves the best PSNR of 19.01 and SSIM of 0.7820, along with the best NIQE score of 9.0, reflecting its effectiveness in both pixel-level accuracy and perceptual quality. CSHNet's LPIPS score of 0.661 also outperforms many other supervised methods, demonstrating its ability to generate perceptually realistic images. FID remains marginally behind SwinG, indicating that SwinG’s Transformer-dominated architecture may slightly outperform CSHNet in capturing global stylistic consistency. Furthermore, CSHNet's RMSE of 23.9 is the lowest among all methods, showcasing its ability to generate highly accurate pixel reconstructions. These results highlight that CSHNet excels in maintaining a balance between structural fidelity and perceptual realism across both datasets. The variations in FID and LPIPS scores suggest that while CSHNet is highly effective in generating high-quality images, some methods, like SwinG, may capture global structure and style slightly better. The discrepancies in LPIPS and FID highlight the inherent trade-offs in optimizing for both local detail preservation and global consistency, which are particularly challenging for hybrid architectures tackling complex image-to-image translation tasks. Finally, CSHNet achieves state-of-the-art performance across most metrics, with minor discrepancies highlighting the inherent challenges of optimizing for both structural fidelity and perceptual realism.

While the proposed CSHNet exhibits robust performance across a range of diverse datasets, its effectiveness in highly complex scenarios, such as multi-modal or multi-spectral environments, presents opportunities for further refinement. Future research could investigate the integration of semantic and multi-modal translation mechanisms, as well as cross-modal feature representations, to enhance its adaptability to heterogeneous datasets. Moreover, leveraging the capabilities of large-scale pre-trained models, holds promise for improving the generalizability of CSHNet in challenging and dynamic settings. By synergizing these advancements with state-of-the-art efficient training strategies, the proposed methodology can be further developed to address the increasing complexity and realism of recent applications.

\subsection{Experimental Comparison of Various Hybrid Networks}
To highlight the advantages of the proposed SCB, we compared it with hybrid networks combining CNN and Transformer architectures, specifically CvT and CMT, using 8 Convolutional Transformer Blocks each for a fair evaluation. Results in Table \ref{tab6} show that SCB outperforms CvT and CMT across most evaluation metrics, except for PSNR, where it slightly trails CMT. Additionally, images generated by SCB, illustrated in Figure \ref{fig8}, demonstrate enhanced detail and contour accuracy, reducing regional blending. Previous studies indicate that large-kernel CNNs, akin to Transformer architectures, provide larger effective receptive fields. We experimented with RepLKNet \cite{ding2022scaling} , a large convolutional kernel of size $31 \times 31$, but found it underperformed compared to SCB due to SCB's superior structural advantages. In conclusion, SCB significantly excels in information asymmetric image translation tasks.

\begin{table}
	\centering
        \setlength\tabcolsep{1.5pt}
	\caption{Evaluation metrics with different hybrid networks.}
	\begin{tabular}{l|cccccc}
		\toprule[1.3pt]
		\diagbox{Component}{Metric}            & PSNR $\uparrow$          & SSIM $\uparrow$         & NIQE $\downarrow$         & FID $\downarrow$          \\ \hline
		CvT         (\textit{ICCV21})       & 14.37                     & 0.1829               & \textsl{34.8}                   & 136.2          \\
		CMT         (\textit{CVPR22})       & \textbf{15.0}          & \textsl{0.2110}          & 36.9                            & 147.4 \\
		RepLKNet   (\textit{CVPR22})        & 14.79                 & 0.2029                   & 35.0                         & \textsl{132.0} \\ \hline
		SCB        (\textit{ours})        & \textsl{14.98}         & \textbf{0.2168}          & \textbf{34.3}               & \textbf{91.0}        \\
		\bottomrule[1.3pt]
	\end{tabular}
	
	\label{tab6}
\end{table}

\subsection{Efficiency Comparison of Different Models}

As shown in Tab.\ref{tab7}, we compared the parameters of different methods. It is evident that CSHNet has more parameters than CNN-Based methods, but significantly fewer than Transformer-based methods. Therefore, CSHNet balances the training performance and generative quality between the two approaches.

\begin{table}
    \centering
    \setlength\tabcolsep{1.5pt}
    \caption{Comparison of parameters of different methods.} 
        \setlength{\tabcolsep}{12pt}
    \begin{tabular}{l|c|c}
        \hline
        Method          & Type             & Paras/M \\ \hline
        CSHNet          & Ours             & 30.6    \\
        Pix2PixHD       & CNN-Based        & 25.6    \\
        StegoGAN        & CNN-Based        & 24.5    \\
        SwinG          & Transformer-Based & 83.3    \\
        CvT            & Transformer-Based & 38.9    \\
        CMT            & Transformer-Based & 2078.2  \\ \hline
    \end{tabular}
    \setlength{\abovecaptionskip}{0.1cm}
    \label{tab7}
    \vspace{-0.4cm}
\end{table}

 \section{\textbf{Conclusion}}

This paper proposes a novel hybrid network combining CNN and Swin Transformer (CSHNet) for information asymmetric image translation, aiming to solve the cross-domain information loss and regional blending problems. Within the bottleneck of CSHNet, two modules, SEC and CES, are designed to address detailed and structural biases, respectively. SEC prioritizes CNN-based detail while incorporating structural information from Swin Transformer. Conversely, CES emphasizes the structure of Swin Transformer while complementing detailed information from CNN. The combination of the two facilitates the learning of both detail and structural information by the network. To further enhance the detailed integrity and regional coherence of generated images, two additional components, IGC and AEPL, are introduced. IGC utilizes low-level feature information to guide the image translation process, while AEPL leverages edge information with dynamic thresholds. We compare CSHNet with SOTA image translation methods on scene-level and instance-level datasets. Visualization results and evaluation metrics demonstrate the superiority of CSHNet.

\bibliographystyle{IEEEtran}
\bibliography{./reference}

\begin{IEEEbiography}[{\includegraphics[width=1in,height=1.25in,clip,keepaspectratio]{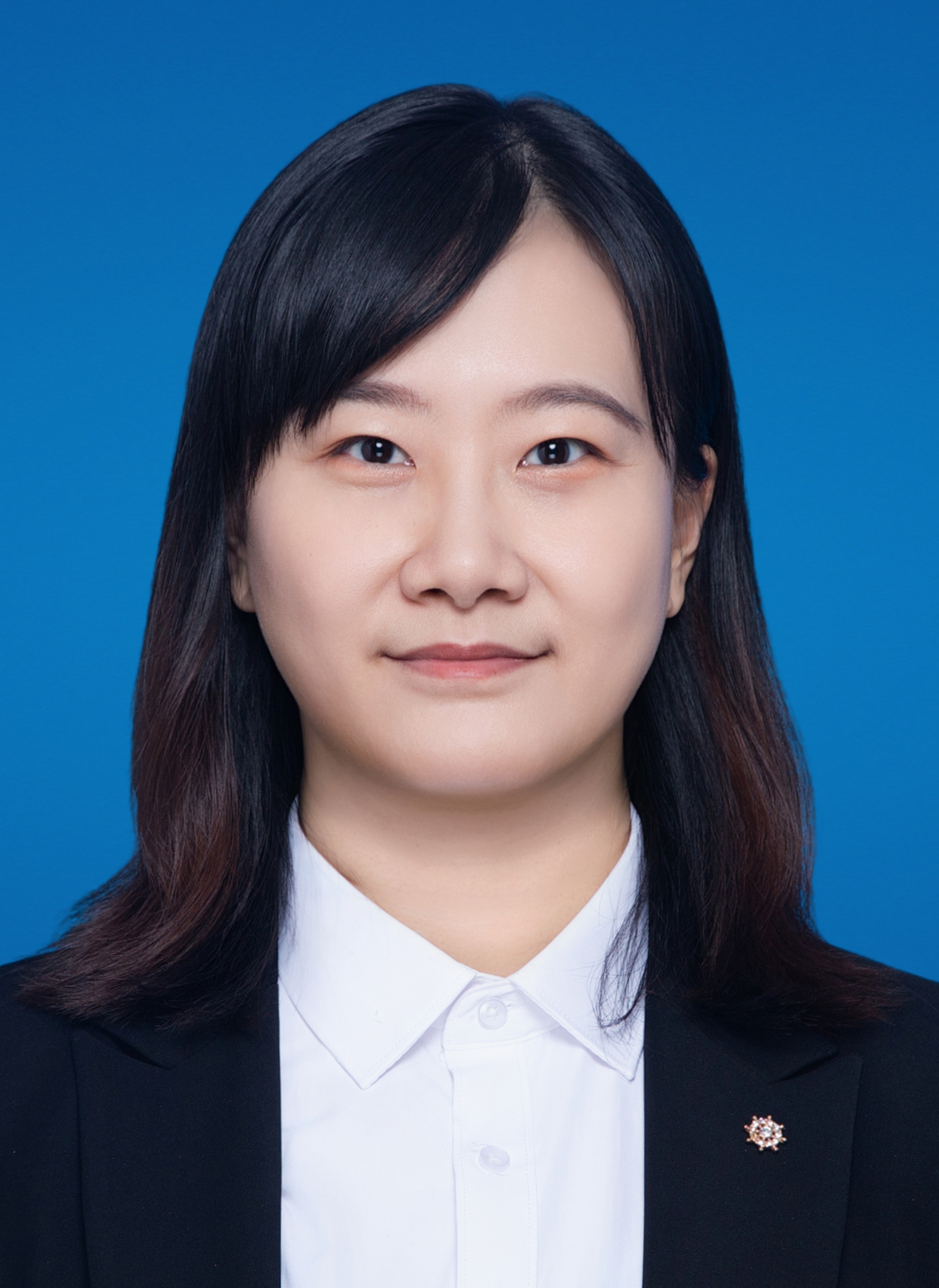}}]{Xi Yang}
	(M'15-SM'24) received the B.Eng. degree in electronic information engineering and the Ph.D. degree in pattern recognition and intelligence system from Xidian University, Xi’an, China, in 2010 and 2015, respectively. From 2013 to 2014, she was a visiting Ph.D. student with the Department of Computer Science, University of Texas at San Antonio, San Antonio, TX, USA. In 2015, she joined the State Key Laboratory of Integrated Services Networks, School of Telecommunications Engineering, Xidian University, where she is currently a Professor in communications and information systems. She has published over 60 articles in refereed journals and proceedings, including IEEE T-TIP, T-NNLS, T-CYB, T-GRS, CVPR, ICCV, ACM MM, etc. Her current research interests include image/video processing, computer vision, and machine learning.
\end{IEEEbiography}

\begin{IEEEbiography}[{\includegraphics[width=1in,height=1.25in,clip,keepaspectratio]{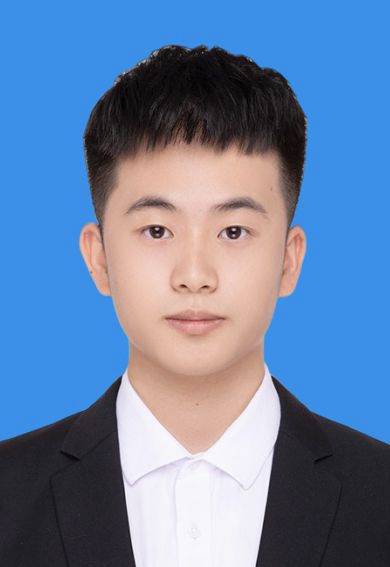}}]{Haoyuan Shi}
	received the B.Eng degree in communication engineering from Hangzhou Dianzi University, Hangzhou, China, in 2023. He is currently pursuing the M.S. degree with electronic and information engineering from Hangzhou Institute of Technology, Xidian University, Hangzhou, China. His current research interests include deep learning and image translation.
\end{IEEEbiography}

\begin{IEEEbiography}
[{\includegraphics[width=1in,height=1.25in,clip,keepaspectratio]
{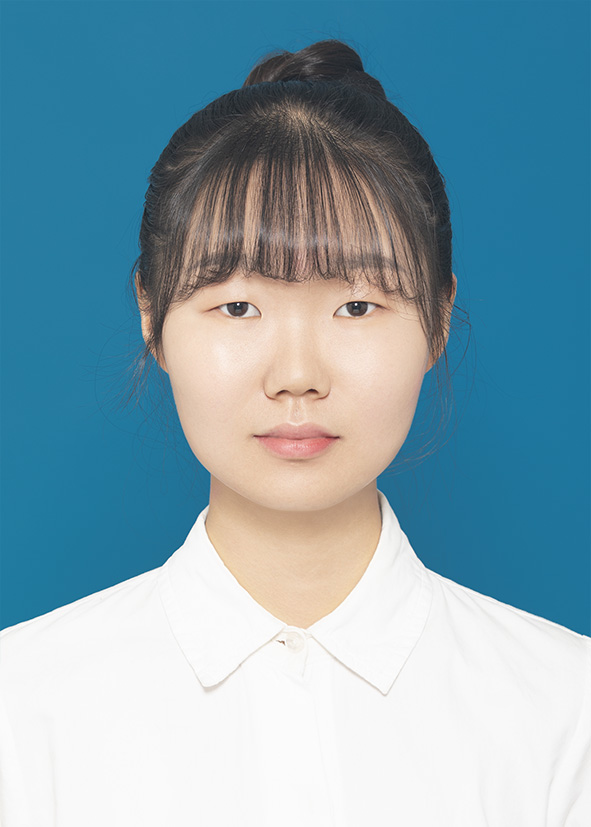}}]{Zihan Wang}
	received the B.Eng. degree in communication engineering from Hebei University of Technology, Hebei, China, in 2021, and the M.S. degree in electronic and communication Engineering from Xidian University, Xi’an, China, in 2024. Her research interests include deep learning and image translation.
\end{IEEEbiography}

\begin{IEEEbiography}[{\includegraphics[width=1in,height=1.25in,clip,keepaspectratio]{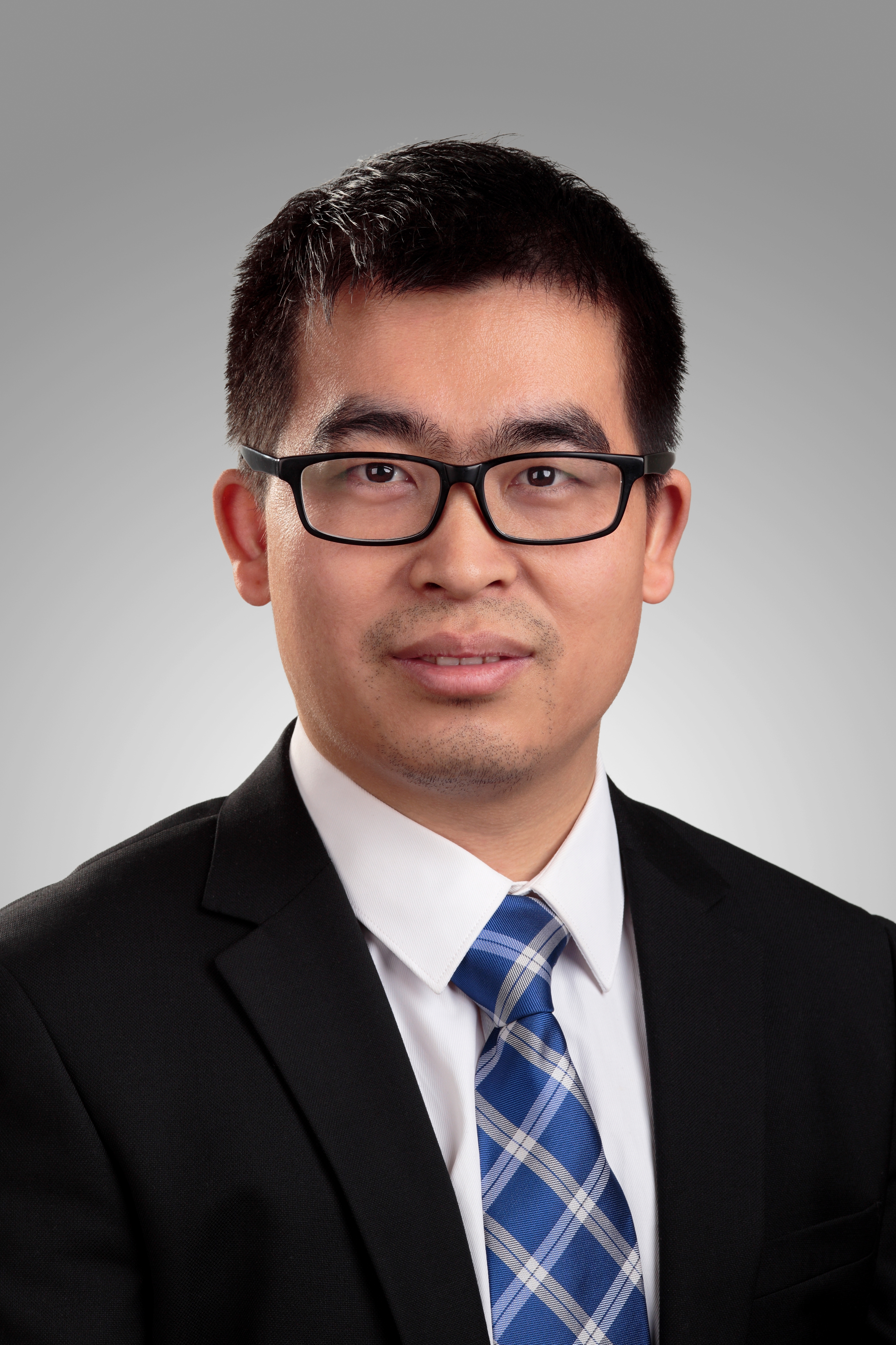}}]{Nannan Wang}
	(M'16-SM'24) received the B.Sc. degree in information and computation science from the Xi'an University of Posts and Telecommunications in 2009 and the Ph.D. degree in information and telecommunications engineering from Xidian University in 2015. He is currently a Professor with the State Key Laboratory of Integrated Services Networks, Xidian University. He has published over 150  articles in refereed journals and proceedings, including IEEE T-PAMI, IJCV, CVPR, ICCV etc. His current research interests include computer vision and machine learning.
\end{IEEEbiography}

\begin{IEEEbiography}[{\includegraphics[width=1in,height=1.25in,clip,keepaspectratio]{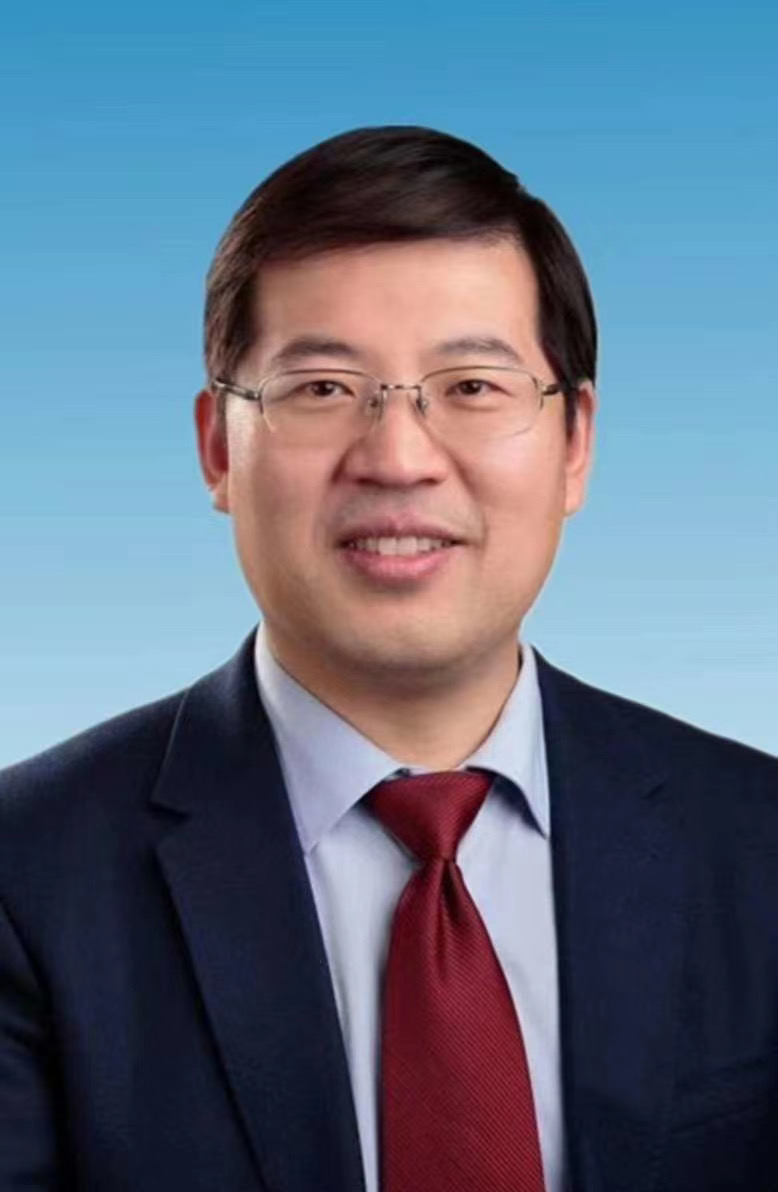}}]{Xinbo Gao}
	(M'02-SM'07-F'24) received the B.Eng., M.Sc. and Ph.D. degrees in electronic engineering, signal and information processing from Xidian University, Xi'an, China, in 1994, 1997, and 1999, respectively. From 1997 to 1998, he was a research fellow at the Department of Computer Science, Shizuoka University, Shizuoka, Japan. From 2000 to 2001, he was a post-doctoral research fellow at the Department of Information Engineering, the Chinese University of Hong Kong, Hong Kong. Since 1999, he has been at the School of Electronic Engineering, Xidian University and now he is a Professor of Pattern Recognition and Intelligent System of Xidian University. Since 2020, he has been also a Professor of Computer Science and Technology of Chongqing University of Posts and Telecommunications. His current research interests include computer vision, machine learning and pattern recognition. He has published seven books and around 300 technical articles in refereed journals and proceedings. Prof. Gao is on the Editorial Boards of several journals, including Signal Processing (Elsevier) and Neurocomputing (Elsevier). He served as the General Chair/Co-Chair, Program Committee Chair/Co-Chair, or PC Member for around 30 major international conferences. He is Fellows of IEEE, IET, AAIA, CIE, CCF, and CAAI.
\end{IEEEbiography}

\end{document}